\documentclass{article}

    \PassOptionsToPackage{numbers, compress}{natbib}

\usepackage[final]{neurips_2019}

\usepackage[draft,inline,nomargin,index]{fixme}
\fxsetup{theme=color,mode=multiuser,inlineface=\itshape,envface=\itshape}
\FXRegisterAuthor{ds}{asv}{\colorbox{gray!10!white}{\color{black}Dylan}}
\FXRegisterAuthor{sf}{asf}{\colorbox{blue!10!white}{\color{black}Sorelle}}
\FXRegisterAuthor{eg}{acs}{\colorbox{red!10!white}{\color{black}Emile}}

\newcommand{\predictortheta}{f_\theta\xspace}

\newcommand{\task}{$\mathcal{T}$\xspace}
\newcommand{\dataset}{$\mathcal{D}$\xspace}

\newcommand{\fairregterm}{$\mathcal{R}$\xspace}

\newcommand{\inp}{\mathbf{x}}
\newcommand{\target}{\mathbf{y}}
\newcommand{\learner}{f}
\newcommand{\lossi}{\mathcal{L}_{\mathcal{T}_i}}

\usepackage[utf8]{inputenc} 
\usepackage[T1]{fontenc}    
\usepackage{hyperref}       
\usepackage{url}            
\usepackage{booktabs}       
\usepackage{amsfonts}       
\usepackage{nicefrac}       
\usepackage{microtype}      
\usepackage{graphicx}
\usepackage{xspace}
\usepackage{amssymb}
\usepackage{amsmath}
\usepackage{graphicx}
\usepackage{algorithmic}
\usepackage{float}
\usepackage{subfig}
\usepackage[]{algorithm2e}

\title{Fair Meta-Learning: Learning How to Learn Fairly\thanks{This research was funded in part by the NSF under grant IIS-1633387.  The Titan Xp used for this research was donated by the NVIDIA Corporation.}}

\author{%
  Dylan Slack \thanks{University of California Irvine, Department of Computer Science, work done while an undergraduate at Haverford College}  \\
   \And
  Sorelle Friedler \thanks{Haverford College, Department of Computer Science}
  \And
  Emile Givental \footnotemark[2]
}

\begin{document}

\maketitle

\begin{abstract}
Data sets for fairness relevant tasks can lack examples or be biased according to a specific label in a sensitive attribute. We demonstrate the usefulness of weight based meta-learning approaches in such situations.  For models that can be trained through gradient descent, we demonstrate that there are some parameter configurations that allow models to be optimized from a few number of gradient steps and with minimal data  \textit{which are both fair and accurate}.  To learn such weight sets, we adapt the popular MAML algorithm to \textit{Fair-MAML} by the inclusion of a fairness regularization term.  In practice, Fair-MAML allows practitioners to train fair machine learning models from only a few examples when data from related tasks is available.  We empirically exhibit the value of this technique by comparing to relevant baselines.
\end{abstract}

\section{Introduction}
\label{sec:introduction}

Advances in the field of meta-learning provide methods to train machine learning models that can better generalize to new tasks using previous experiences.  A known issue in developing fair machine learning classifiers is data collection.  Data can be biased in collection, have minimal training examples, or otherwise be unrepresentative of the true testing population \cite{Kallus2018ResidualUI, whatdopracneed, coston2019fair}.  Can we adapt meta-learning approaches to handle issues in fairness related to minimal data and bias in the distribution of training data when there is related task data  available?

We note that overall the problem of training fair machine learning models with very little task specific training data is relatively unstudied.  Related work includes methods to transfer fair machine models.  Madras et. al. propose an adversarial learning approach called LAFTR for fair transfer learning \cite{madras18}.  Schumman et. al. provide theoretical guarantees surrounding transfer fairness related to equalized odds and opportunity and suggest another adversarial approach aimed at transferring into new domains with different sensitive attributes \cite{transferfairnessschumann19}.  Additionally,  Lan and Huan observe that the predictive accuracy of transfer learning across domains can be improved at the cost of fairness \cite{Lan2017DiscriminatoryT}.  Related to fair transfer learning, Dwork et. al. use a decoupled classifier technique to train a selection of classifiers fairly for each sensitive group in a data set \cite{dwork2018decoupled}.   Developing models that are able to achieve satisfactory levels of fairness and accuracy with only minimal data available in the desired task could be an important avenue for future work. To address the proposed question, we introduce a fair meta-learning approach: \textit{Fair-MAML}. 


\section{Fairness Setting}
\label{subsec:fairness}

We assume a binary fair classification scenario.  We have features $X\in \mathbb{R}^n$, labels $Y\in \{0,1\}$ and sensitive attributes $A \in \{0,1\}$.  We consider $1$ the positive outcome in $Y$ and $0$ the protected group in $A$.  We train a classifier parameterized by $\theta$, $f_\theta$ to be accurate and fair with respect to $Y$ and $A$.

We consider two notions of fairness \textit{demographic parity} and \textit{equal opportunity}.  Demographic parity (or disparate impact \cite{feldman2015certifying}) can be described as:

\begin{equation}
\label{eq:demographicparity}
\frac{P(\hat{Y}=1|A=0)}{P(\hat{Y}=1|A=1)}
\end{equation}

Here, we present demographic parity \cite{dwork12} as a ratio. If the ratio is closer to $1$, it indicates more fairness.  Next, equal opportunity \cite{Hardt2016Equality} requires that the protected groups have equivalent \textit{true positive rates}.  Similarly, a value closer to $1$ indicates high levels of fairness:

\begin{equation}
\label{eq:equalizedodds}
\frac{P(\hat{Y}=1|A=0,Y=1)}{P(\hat{Y}=1|A=1,Y=1)} \mbox{~~~~~} 
\end{equation}

Another oft-noted notion of fairness \textit{equalized odds} \cite{Hardt2016Equality} also includes the \textit{false positive rate}.  In this work however, we only consider the two earlier mentioned definitions of group fairness.

\section{Fair-MAML}

\subsection{Fair-MAML Algorithm}





In the meta-learning scenario used in this paper, we train $\predictortheta$ to learn a new task drawn over a distribution of tasks $\mathcal{T}\sim P(\mathcal{T})$ using $K$ examples drawn from \task.  Additionally, we assume $\predictortheta$ can be optimized through gradient descent.   We define a task as $\mathcal{T} = \{\mathcal{D},\mathcal{L},\mathcal{R},\gamma \}$.  Each task includes a fairness regularization term \fairregterm and fairness hyperparameter $\gamma$.  Additionally, it has a dataset \dataset consisting of features $X$, labels $Y$, and sensitive attribute $A$ such that $\mathcal{D}=(X,Y,A)$ as well as loss function $\mathcal{L}$.  


In order to train a fair meta-learning model, we adapt Model-Agnostic Meta-Learning or MAML \cite{Finn2017ModelAgnosticMF} to our fair meta-learning framework and introduce \textit{Fair-MAML} by including the regularization term $\mathcal{R}$ and fairness hyperparameter $\gamma$.  MAML is trained by optimizing performance of $\predictortheta$ across a variety of tasks after one gradient step.  The Fair-MAML algorithm is given in algorithm \ref{alg:fairmaml}.

\begin{algorithm}
\caption{Fair-MAML}
\label{alg:fairmaml}
\begin{algorithmic}
{
\REQUIRE $p(\mathcal{T})$: distribution over tasks
\REQUIRE $\alpha$, $\beta$: step size hyperparameters
\STATE randomly initialize $\theta$
\WHILE{not done}
\STATE Sample batch of tasks $\mathcal{T}_i \sim p(\mathcal{T})$
  \FORALL{$\mathcal{T}_i$}
      \STATE Sample $K$ datapoints $\mathcal{D}=\{\inp^{(j)}, \target^{(j)}, \textbf{a}^{(j)}$\} from $\mathcal{T}_i$
      \STATE Evaluate $\nabla_\theta \lossi(\learner_\theta)$ using $\mathcal{D}$ and $\lossi$ 
      \STATE Compute updated parameters: $\theta_i'=\theta-\alpha \nabla_\theta [ \lossi(  \learner_\theta ) + \gamma_{\mathcal{T}_i} \mathcal{R}_{\mathcal{T}_i}(\learner_\theta)]$
      \STATE Sample $K$ \textit{new} datapoints $\mathcal{D}_i'=\{\inp^{(j)}, \target^{(j)}, \textbf{a}^{(j)} \}$ from $\mathcal{T}_i$ to be used in the meta-update
 \ENDFOR
 \STATE Update $\theta \leftarrow \theta - \beta \nabla_\theta \sum_{\mathcal{T}_i \sim p(\mathcal{T})} [ \lossi ( \learner_{\theta_i'}) + \gamma_{\mathcal{T}_i} \mathcal{R}_{\mathcal{T}_i}(\learner_{\theta_i'})]$ using each $\mathcal{D}_i'$ 
\ENDWHILE
}
\end{algorithmic}
\end{algorithm}


\subsection{Fairness Regularizers}
Fair-MAML requires that second derivatives be computed through a Hessian-vector product in order to calculate the meta-loss function which can be computationally intensive and time-consuming. We propose two regularization terms for demographic parity and equal opportunity that are quick to compute.  First considering demographic parity, let $\mathcal{D}_0$ denote the protected instances in $X$ and $Y$:

\begin{equation}
        \mathcal{R}_{dp}(\predictortheta,\mathcal{D}) = 1 - P(\hat{Y}=1|A=0) \approx  1 - \frac{1}{|\mathcal{D}_0|}\sum_{x \in \mathcal{D}_0} P(f_\theta(x) = 1)
   \label{eq:di_reg}
\end{equation}
%
This regularizer incurs penalty if the probability that the protected group receives positive outcomes is low. Our value assumption is that we attempt to adjust upwards the rate at which the protected class receives positive outcomes. 

We also propose a regularizer for equal opportunity. Let $\mathcal{D}_0^1$ denote the instances within $X$ that are both protected and have the positive outcome in $Y$.

\begin{equation}
        \mathcal{R}_{eop}(\predictortheta,\mathcal{D}) = 1 - P(\hat{Y}=1|A=0,Y=1) 
         \approx  1 - \frac{1}{|\mathcal{D}_0^1|}\sum_{x \in \mathcal{D}_0^1} P(f_\theta(x) = 1)
   \label{eq:equal_opp_reg}
\end{equation}

We have a similar value assumption for equal opportunity.  We adjust the true positive rate of the protected class upwards.


\section{Experiments}

\subsection{Synthetic Experiment}
\label{sec:synthetic_experiment}

We illustrate the usefulness of Fair-MAML as opposed to a regularized pre-trained model through a synthetic example based on Zafar et. al \cite{zafar17}.
We generate two Gaussian distributions using the means and covariances from Zafar et. al.  To simulate a variety of tasks, we generate $y$ labels by letting all the positively labeled points be those above a randomly generated line through $(0,0)$ with slope randomly selected on the range $[-5,5]$.  We add a sensitive feature using the same technique from Zafar et. al.  Full data generation detail can be found in section \ref{sec:datagendeets} in the appendix.

In order to assess the fine-tuning capacity of Fair-MAML and the pre-trained neural network, we introduce a more difficult fine-tuning task.  During training, the two classes were separated clearly by a line. For fine-tuning, we set each of the binary class labels to a specific distribution.  
In this scenario, a straight line cannot clearly divide the two classes.  We assigned sensitive attributes using the same strategy as above.  Additionally, we only fine-tuned with $5$ \textit{positive-outcome} examples from the \textit{protected class}.  

We trained Fair-MAML using a neural network consisting of two hidden layers of $20$ nodes and the ReLU activation function. Full training details can be found in \ref{sec:trainingdeets} in the appendix. We present the biased example task in figure \ref{fig:synthetic_example}. 
We give comprehensive results over a variety of examples in the appendix in figure \ref{fig:more_synthetic_examples}.
In the new task, there is an unseen configuration of positively labeled points. 
It was not possible for positively labeled points to fall below $y=0$ during training. Fair-MAML is able to perform well with respect to both fairness and accuracy on the fine-tuning task when only biased fine-tuning data is available while the pre-trained network fails.  


\bgroup
\begin{figure*}

\includegraphics[scale=.17]{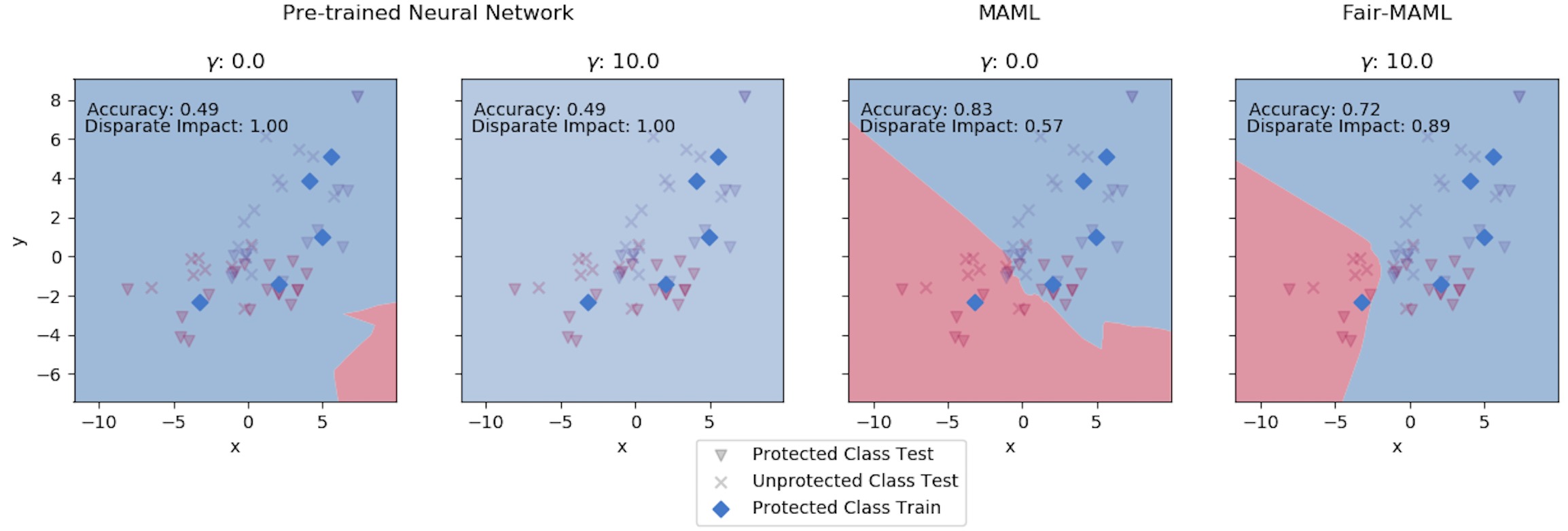}

\caption{An example decision boundary from the pre-trained neural network, MAML, and Fair-MAML on the synthetic example (note: Fair-MAML is MAML with $\gamma=0$).  Points that are colored the same as the side of the boundary are correct.  Blue is the positive class.  Only points in the positive outcome and protected class are given for the fine-tuning task.  Fair-MAML is able to handle such an imbalance of training points on a previously unseen task while the pre-trained neural network fails---illustrating that Fair-MAML has learned a more useful internal representation.}
\label{fig:synthetic_example}
\end{figure*}
\egroup

\subsection{Communities and Crime Experiment}
\label{subsubsec:ccexp}

Next we consider an example using the Communities and Crime data set \cite{communites_and_crime}.  
The goal is to predict the violent crime rate from community demographic and crime relevant information.  
We convert this data set to a multi-task format by using each state as a different task.  
We convert violent crime rate into a binary label by whether the community is in the top $50\%$  of violent crime rate within a state.  
We add a binary sensitive column for whether African-Americans are the highest or second highest population in a community in terms of percentage racial makeup.  

We trained two Fair-MAML models---one with the demographic parity regularizer from equation \ref{eq:di_reg} and another with the equal opportunity regularizer from equation \ref{eq:equal_opp_reg}.  We used a neural network with two hidden layers of $20$ nodes.  
Additionally, we trained two transfer LAFTR models \cite{madras18} on the transfer tasks using the demographic parity and equal opportunity adversarial objectives.  We also trained a pre-trained network regularized for both demographic parity and equal opportunity.  We set $K$=10 in Fair-MAML and fine-tuned on $10$ communities for both Fair-MAML and the pre-trained network.  We used the fairness regularizers for both Fair-MAML and the pre-trained network while fine-tuning. Full training details can be found in section \ref{sec:ccdeets} in the appendix.

\begin{figure*}

\includegraphics[scale=.16]{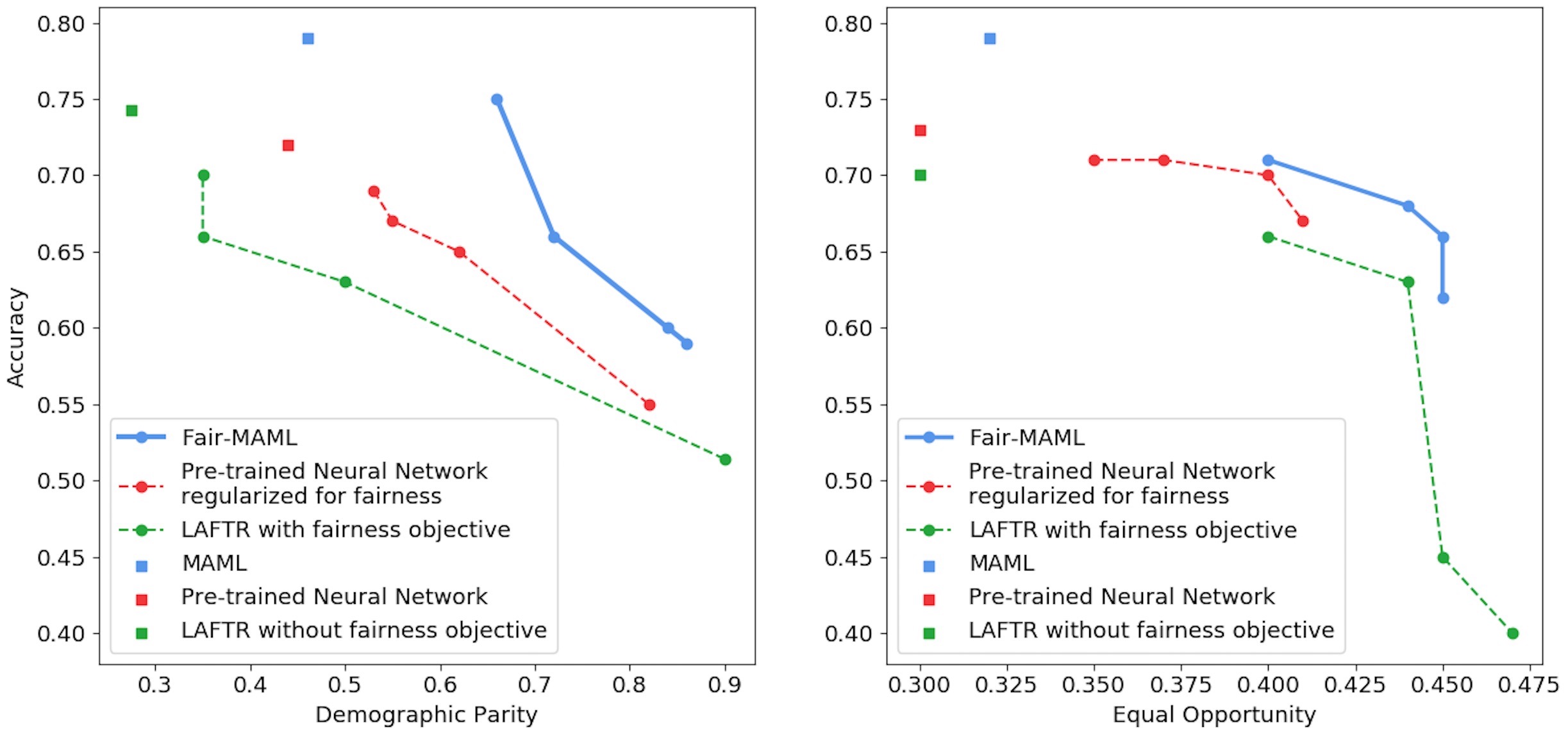}

\caption{The accuracy/fairness trade off for the communities and crimes example sweeping over a range of $\gamma$'s.  The data presented is the mean across three runs on each $\gamma$ using $5$ randomly selected hold out tasks.  
Higher accuracy and fairness values closer to $1.0$ indicate more successful outcomes.  The pre-trained neural network and Fair-MAML received $10$ fine-tuning points and were optimized for $1$ epoch.  
We did not find useful results using LAFTR with only $10$ fine-tuning points or with a minimal number of fine-tuning epochs, so the LAFTR example given here is with $30$ fine-tuning points and $100$ epochs of optimization.  
Fair-MAML scores the strongest levels of accuracy and fairness.
}
\label{fig:cc_results}
\end{figure*}

When training a MLP from the encoder on each of the transfer tasks, we found that LAFTR struggled to produce useful results with only $10$ training points from the new task over any number of training epochs.  We found that we were able to get reasonable results from LAFTR using $30$ fine-tuning points and $100$ epochs of optimization.  It makes sense that a smaller number of training epochs for the new task is unsuccessful because the MLP trained on the fairly encoded data is trained from scratch.  The results are presented in figure \ref{fig:cc_results}.  Fair-MAML is able to achieve the best balance of fairness and accuracy and demonstrates strong ability to generalize to new states using only $10$ communities.

\section{Conclusions}

We propose studying transfer fairness in situations with minimal task specific training data.  We showed the usefulness of weight-based meta-learning through the introduction of Fair-MAML to accurately/fairly train models on both synthetic and real data sets from only a couple training points.  We leave evaluating other meta-learning approaches including first order MAML to future work.

\bibliographystyle{plain}
\bibliography{bib}

\section{Appendix}

\begin{figure}[h]
\centering
    \includegraphics[width=1.02\textwidth]{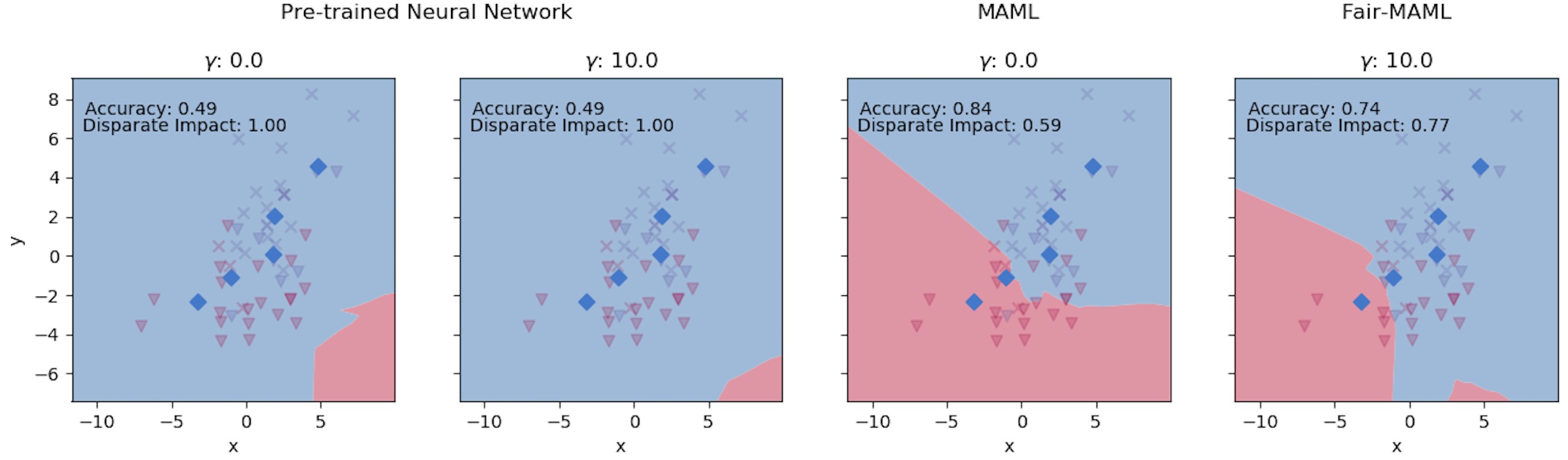}
    \includegraphics[width=1.02\textwidth]{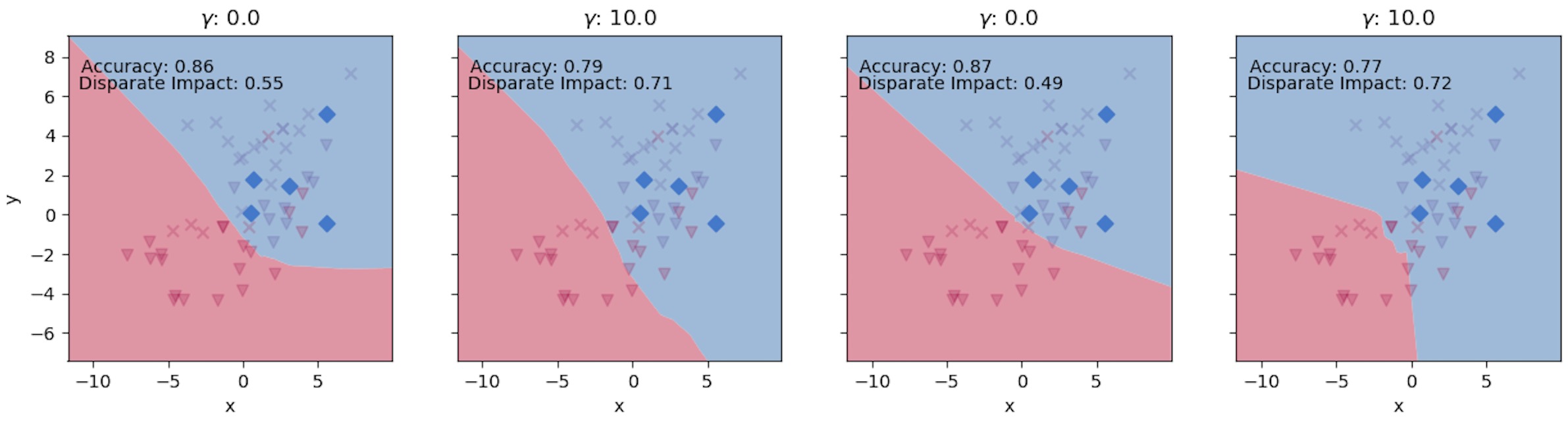}
    \includegraphics[width=1.02\textwidth]{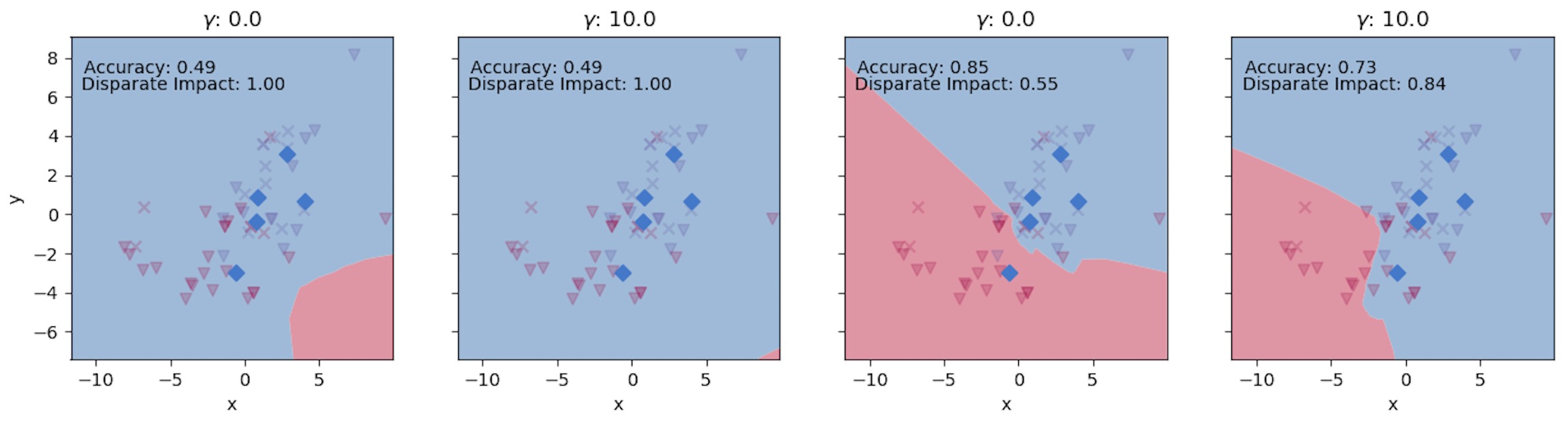}
    \includegraphics[width=1.02\textwidth]{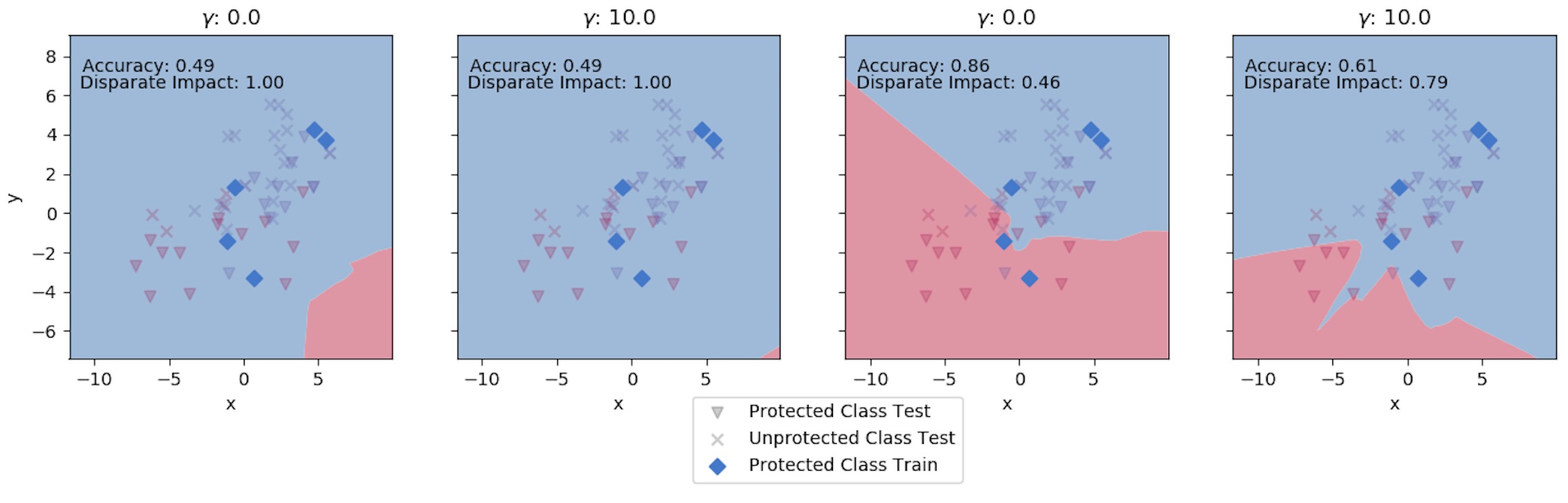}
    \caption{Additional synthetic examples.}
\label{fig:more_synthetic_examples}
\end{figure}

\subsection{Data Generation Synthetic Task}
\label{sec:datagendeets}

The first distribution (1) is set to $p(x) = N([2; 2], [5, 1; 1, 5])$ and the second (2) is set to $p(x) = N([-2; -2], [10, 1; 1, 3])$.  During training, we simulate a variety of tasks by dividing the class labels along a line with y-intercept of $(0,0)$ and a slope randomly selected on the range $[-5,5]$.  All points above the line in terms of their $y$-coordinate receive a positive outcome while those below are negative.  Using the formulation from Zafar et. al., we create a sensitive feature by drawing from a Bernoulli distribution where the probability of the example being in the protected class is: $p(a=0)= p(x'|y=1)/(p(x'|y=1)+p(x'|y=0))$ where $x' = [cos(\phi), -sin(\phi);sin(\phi), cos(\phi))]x$.  Here, $\phi$ controls the correlation between the sensitive attribute and class labels.  The lower $\phi$, the more correlation and unfairness.  We randomly select $\phi$ from the range $[2,4,8,16]$ to simulate a variety in fairness between tasks.

\subsection{Training Details Synthetic Task}
\label{sec:trainingdeets}
We randomly generated $100$ synthetic tasks that we cached before training.  We sampled $5$ examples from each task during meta-training, used a meta-batch size of $32$ for Fair-MAML, and performed a single epoch of optimization within the internal MAML loop. We trained Fair-MAML for $5,000$ meta-iterations.  For the pre-trained neural network, we performed a single epoch of optimization for each task.  We trained over $5,000$ batches of $32$ tasks per batch to match the training set size used by Fair-MAML.  

The loss used is the cross-entropy loss between the prediction $f(x)$ and the true value using the demographic parity regularizer from equation \ref{eq:di_reg}.  We use a neural network with two hidden layers consisting of $20$ nodes and the ReLU activation function.  We used the softmax activation function on the last layer.  When training with Fair-MAML, we used $K=5$ examples and performed one gradient step update.  We set the step size $\alpha$ to $0.3$, used the Adam optimizer to update the meta-loss with learning rate $\beta$ set to $1e-3$.  We pre-trained a baseline neural network on the same architecture as Fair-MAML.  To one-shot update the pre-trained neural network we experimented with step sizes of $[0.01,0.1,0.2,0.3]$ and ultimately found that $0.3$ yielded the best trade offs between accuracy and fairness.  Additionally, we tested $\gamma$ values during training and fine-tuning of $[0,10]$.  

\subsection{Communities and Crime Experiment Details}
\label{sec:ccdeets}

\subsubsection{Additional pre-processing details}
The Communities and Crime data set has data from $46$ states ranging in number of communities from $1$ to $278$ communities per state.  We only used states with $20$ or more communities leaving $30$ states.  We held out $5$ randomly selected states for testing and trained using $25$ states.    

\subsubsection{Fair-MAML Training Details}

We set $K=10$ and cached $100$ meta-batches of size $8$ states for training.  For testing, we randomly selected $10$ communities from the hold out task that we used for fine-tuning and evaluated on whatever number of communities were left over.  The number of evaluation communities is guaranteed to be at least $10$ because we only included states with $20$ or more communities. 

Each of the two layers of $20$ nodes used the ReLU activation function.  We trained Fair-MAML with one gradient step using a step size of $1e-2$ and a meta-learning rate of $1e-4$ using the Adam optimizer.  We trained the model for $2,000$ meta-iterations.  
In order to assess Fair-MAML, we trained a neural network regularized for fairness using the same architecture and training data.  We fine-tuned the neural network for each of the assessment tasks.  We used a learning rate of $1e-3$ for training and assessed learning rates of $[1e-4,1e-3,1e-2,1e-1]$ for fine-tuning.  We found the fine-tuning rate of $1e-1$ to perform the best trade offs between accuracy and fairness and present results using this learning rate.  We varied $\gamma$ over $[0,4]$ incremented by $1$ for the demographic parity regularizer.  We found higher $\gamma$'s to work better for the equal opportunity regularizer and varied $\gamma$ from $[0,40]$ incremented by $10$.

\subsubsection{LAFTR Training Details}

 We used the same transfer methodology and hyperparameters as described in Madras et. al. \cite{madras18} and used a neural network with a hidden layer of $20$ nodes as the encoder.  We used another neural network with a hidden layer of $20$ nodes as the MLP to be trained on the fairly encoded representation.  We used the demographic parity and equal opportunity adversarial objectives for the first and second LAFTR model respectively.  

We trained each encoder for $1,000$ epochs and swept over a range of $\gamma's$: $[0,0.5,1.0,2.0,4.0]$.  We trained with all the data not held out as one of the $5$ testing tasks. Though we do not include the results in presentation, we were able to generate similar results with LAFTR to Fair-MAML using $50$ training points from the new task after $100$ epochs of optimization.

\end{document}